\documentclass[10pt,twocolumn,letterpaper]{article}

\usepackage{iccv}              

\PassOptionsToPackage{table, dvipsnames}{xcolor}
\usepackage{xcolor}
\usepackage{graphicx}
\usepackage{amsmath}
\usepackage{amssymb}
\usepackage{booktabs}
\usepackage{stfloats}
\usepackage[accsupp]{axessibility}

\definecolor{iccvblue}{rgb}{0.21,0.49,0.74}
\usepackage[pagebackref,breaklinks,colorlinks,allcolors=iccvblue]{hyperref}

\usepackage[capitalize]{cleveref}
\crefname{section}{Sec.}{Secs.}
\Crefname{section}{Section}{Sections}
\Crefname{table}{Table}{Tables}
\crefname{table}{Tab.}{Tabs.}

\usepackage{multirow}
\usepackage{listings}
\usepackage{bbm}
\usepackage{amsmath}
\usepackage{amssymb,algorithm}
\usepackage{algpseudocode}
\usepackage{amssymb}
\usepackage{pifont}

\usepackage[normalem]{ulem}
\usepackage{colortbl}

\definecolor{codegreen}{rgb}{0,0.6,0}
\definecolor{codegray}{rgb}{0.5,0.5,0.5}
\definecolor{codepurple}{rgb}{0.58,0,0.82}
\definecolor{backcolour}{rgb}{0.95,0.95,0.92}

\newlength\savewidth

\lstdefinestyle{mystyle}{
    commentstyle=\color{codegreen},
    keywordstyle=\color{blue},
    stringstyle=\color{codepurple},
    basicstyle=\ttfamily\footnotesize,
    breakatwhitespace=false,         
    breaklines=true,                 
    captionpos=b,                    
    keepspaces=true,                 
    showspaces=false,                
    showstringspaces=false,
    showtabs=false,                  
    tabsize=2
}
\def\ours{\textsc{GTA-CLIP}}

\newcommand{\A}{\mathcal{A}}
\newcommand{\X}{\mathcal{X}}
\newcommand{\Y}{\mathcal{Y}}

\newcommand{\bmu}{\boldsymbol{\mu}}
\newcommand{\bS}{\boldsymbol{\Sigma}}
\newcommand{\bx}{\mathbf{x}}
\newcommand{\by}{\mathbf{y}}
\newcommand{\bz}{\mathbf{z}}
\newcommand{\ba}{\mathbf{a}}
\newcommand{\bp}{\mathbf{p}}

\DeclareMathOperator{\argmax}{argmax}

\def\paperID{10269} 
\def\confName{ICCV}
\def\confYear{2025}

\title{Generate, Transduct, Adapt: Iterative Transduction with VLMs}

\author{Oindrila Saha \quad \quad Logan Lawrence \quad \quad Grant Van Horn \quad \quad  Subhransu Maji\\
University of Massachusetts, Amherst\\
{\tt\small \{osaha, lclawrence, gvanhorn, smaji\}@umass.edu}}

\begin{document}

\twocolumn[{%
\renewcommand\twocolumn[1][]{#1}%
\maketitle
\begin{center}
    \centering
    \captionsetup{type=figure} 
    \includegraphics[width=0.95\linewidth]{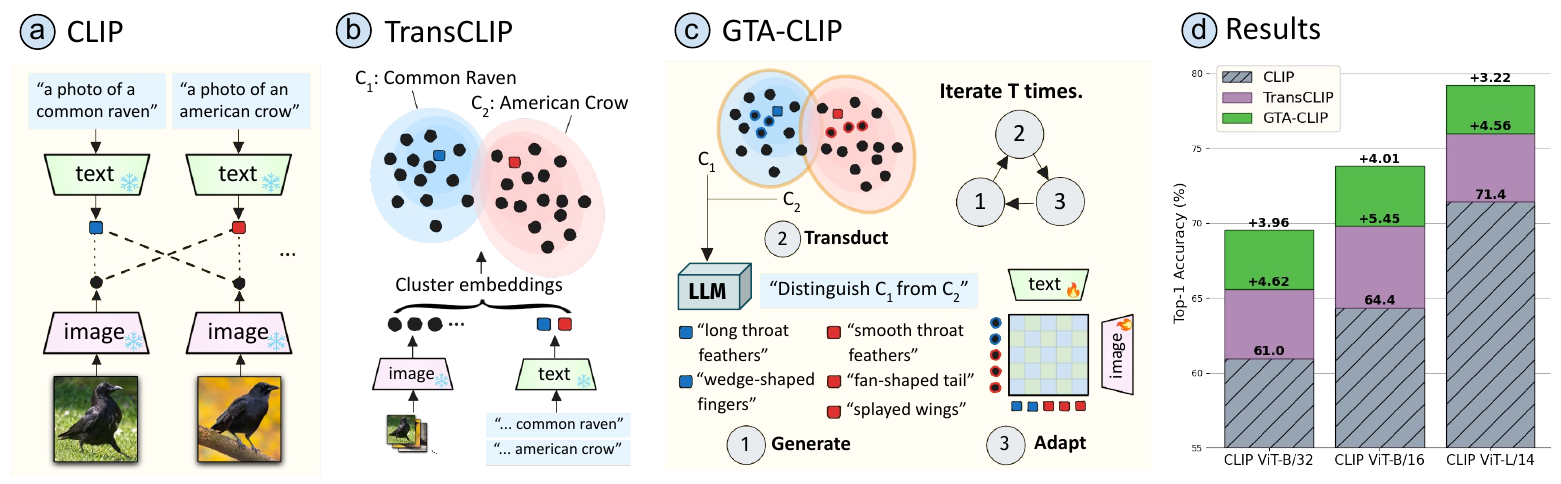}
    \vspace{-10pt}
    
    \captionof{figure}{\textbf{Overview of \ours{}.}~\textbf{(a)} Vision-language models (VLMs) such as CLIP~\cite{clip} enable zero-shot classification using similarity between text embeddings of class prompts and images.~\textbf{(b)}  Transductive CLIP~\cite{transclip} exploits the structure of the entire image dataset to assign images to classes improving accuracy.~\textbf{(c)} Our approach, \ours{}, iteratively (i) induces structure over the classes in language space by generating attributes driven by the pairwise confusions, (ii)  performing attribute-augmented transductive inference, and (iii) adapting CLIP encoders using the inferred labels. \textbf{(d)} Across 12 datasets we improve upon CLIP and transductive CLIP by 9.5\% and 4.0\% using VIT-B/16, and similarly for other encoders. Significant improvements are also reported in the few-shot setting. 
    }
    \label{fig:motivation}
   \vspace{-1mm}
\end{center}%
}]

\begin{abstract}
Transductive zero-shot learning with vision-language models leverages image-image similarities within the dataset to achieve better classification accuracy compared to the inductive setting. 
However, there is little work that explores the structure of the language space in this context.
We propose \ours{}, a novel technique that incorporates supervision from language models for joint transduction in language and vision spaces.
Our approach is iterative and consists of three steps: (i) incrementally exploring the attribute space by querying language models, (ii) an attribute-augmented transductive inference procedure, and (iii) fine-tuning the language and vision encoders based on inferred labels within the dataset.
Through experiments with CLIP encoders, we demonstrate that \ours{} yields an average performance improvement of 9.5\% and 4.0\% across 12 datasets and 3 encoders, over CLIP and transductive CLIP respectively in the zero-shot setting. 
We also observe similar improvements in a few-shot setting.  
We present ablation studies that demonstrate the value of each step and visualize how the vision and language spaces evolve over iterations driven by the transductive learning.
\end{abstract}
\vspace{-12pt}
\section{Introduction}
\label{sec:intro}
Recent advances in vision-language models (VLMs) have enabled zero-shot image classification across diverse domains. These models, such as CLIP~\cite{clip}, assign images to classes based on the similarity between image and text embeddings, forming the basis of various zero-shot approaches in classification~\cite{cocoop, coop, tip_adapter}, segmentation~\cite{xu2022simple, rao2022denseclip, luddecke2022image, luo2023segclip}, and detection~\cite{minderer2024scaling, zhong2022regionclip} (Fig.~\ref{fig:motivation}a). 
However, in many practical scenarios, the images requiring classification are known in advance. For example, an ecologist might have a large collection of animal images that need to be categorized by species. In such cases, \textit{transductive inference} is more suitable, as it leverages the dataset’s inherent structure to refine predictions (Fig.\ref{fig:motivation}b).

Despite the success of transductive inference with VLMs, existing approaches often overlook the rich structure of the label space derived from language. For instance, linking semantically similar descriptions or attributes can yield more coherent class prototypes, while aligning these attributes with image features can enable model \emph{adaptation} on the specific dataset. This strategy can be advantageous for zero-shot and few-shot recognition, especially in novel or fine-grained domains where labeled data is scarce.

To address this gap, we propose \ours{}, a transductive learning approach that exploits structure in both the language and vision spaces (Fig.\ref{fig:motivation}c and Alg.\ref{algo:jtclip}). Our method begins by querying a language model to populate the language space: starting with an initial set of attributes per category, we dynamically expand this space by \emph{generating} discriminative attributes based on pairwise confusion between classes. This strategy improves class separation while maintaining computational tractability. We then design a \textit{transductive inference} procedure that refines predictions using these attributes. Finally, we \textit{adapt} the underlying VLM to the target dataset using inferred labels and attributes. This iterative cycle of Generation, Transduction, and Adaptation—hence the name GTA—progressively improves recognition performance. 

We present experiments on a benchmark of 12 datasets using various CLIP encoders, where our approach achieves 8.6\% improvement over CLIP and 3.7\% improvement over the current state-of-the-art transductive CLIP~\cite{transclip} on average (Fig.~\ref{fig:motivation}d and Table~\ref{table:full_test_set_performances}). Notably, on a dataset like CUB with about 12k images, the whole process completes in 12-20 minutes on a single A100 GPU (see \S~\ref{sec:computation}).
Ablation studies demonstrate that each component of our method contributes to these gains. Specifically, while attribute-augmented transduction improves performance on average, it is most effective when paired with model fine-tuning. Similarly, dynamically expanding the attribute space benefits fine-grained domains while keeping learning efficient. We visualize how the language and vision spaces evolve over iterations providing insights into the performance improvements. Further, we demonstrate \ours{}’s advantages in a few-shot setting (Table~\ref{table:few_shot_performance}) and a zero-shot setting where labeled examples from related categories are available during training (Table~\ref{table:adaptclipzs_comp}). In both cases, our approach outperforms transductive CLIP~\cite{transclip} and prior methods. 

To summarize, our main contribution is to demonstrate that zero- and few-shot classification can be significantly improved in a transductive setting by integrating attribute generation, transductive inference, and model adaptation into a unified framework. While prior work has explored these components in isolation, to the best of our knowledge, this is the first work to show that their benefits are complementary and can be effectively leveraged in label-scarce scenarios. Our approach is of practical value as it provides end users with another avenue to improve labeling accuracy on their target dataset, alongside traditional labeling efforts. Our code is released at \url{https://github.com/cvl-umass/GTA-CLIP}.

\section{Related Work}
\noindent\textbf{Transductive learning}~\cite{vapnik1998support} is well-suited for scenarios where a model’s predictions must be accurate on a specific dataset rather than on unseen future data. Access to the entire unlabeled test set enables inference through methods such as label propagation~\cite{fu2015transductive,wang2021zero}, clustering~\cite{wan2019transductive, yu2017transductive, wang2021zero}, among others~\cite{bo2021hardness, liu2021iterative}. This setting closely resembles semi-supervised learning, where techniques like pseudo-labeling~\cite{zhang2021flexmatch, cascante2021curriculum, arazo2020pseudo}, entropy minimization~\cite{grandvalet2004semi}, and self-training~\cite{yalniz2019billion, xie2020self} have proven effective.

\noindent\textbf{Zero-shot transduction} has been previously explored using image generation~\cite{gao2020zero, wang2018zero} and attribute-based approaches~\cite{xu2017transductive, yu2018transductive}, while more recent methods leverage VLMs to estimate initial class prototypes from language. For example, ZLaP~\cite{kalantidis2024label} improves CLIP through label propagation, while~\cite{martin2024transductive} iteratively estimate assignments and class prototypes. TransCLIP~\cite{transclip} presents an efficient approach for large-scale zero-shot transduction, employing a block majorization-minimization (BMM) algorithm~\cite{hong2017iteration, razaviyayn2013unified} to optimize an objective comprising: a Gaussian mixture model, a Laplacian regularizer, and a KL divergence term that aligns assignments with image-text probabilities across the dataset. \textit{We extend this state-of-the-art by incorporating class-specific attributes in the KL divergence term, enabling better alignment of image features with the semantic structure of the dataset.}

\noindent\textbf{Improving Zero-shot with Attributes.} Large language models (LLMs) have been used to improve zero-shot classification by expanding attribute spaces beyond simple class names. For example,~\cite{pratt2023does, menon2022visual, naeem2023i2mvformer} employ LLMs to generate rich category descriptions (e.g., describing a tiger as having stripes and claws) to improve both classification accuracy and interpretability in CLIP-based models. Beyond LLM-based augmentation, other approaches focus on identifying a concise and discriminative set of attributes for recognition~\cite{yan2023learning, chiquier2024evolving}. Inspired by both strategies, we leverage language models such as GPT~\cite{achiam2023gpt} and LLaMA~\cite{touvron2023llama} to initially populate the attribute space. \textit{However, rather than relying on static expansions, we introduce a dynamic refinement process: attributes are iteratively added to classes that are frequently confused, improving class separability.}

\noindent\textbf{Adapting CLIP.} Prior work has shown that augmenting CLIP with attributes does not significantly improve zero-shot recognition, particularly in out-of-domain or fine-grained datasets. In such cases, model adaptation is necessary. Existing techniques range from learning language and vision prompts~\cite{coop,cocoop} to incorporating learnable layers~\cite{clip_adapter,tip_adapter} or performing full fine-tuning~\cite{singha2023ad, tian2022vl, wortsman2022robust, zhang2023domain}. A different line of work addresses fine-tuning without paired image and text data. WiSE-FT \cite{wortsman2022robust} and LaFTer \cite{mirza2024lafter} use ensembles, while  others~\cite{lewis2023gist,clipftaself,saha2024} show the value of large-scale fine-tuning with image-text data aligned at the category level. We build on the approach of AdaptCLIPZS~\cite{saha2024}, which stochastically pairs images with attributes within a category and modifies the CLIP objective to accommodate weaker supervision. \textit{However, while all the above approaches rely on labeled examples, such as images with class labels on the target domain, our method enables adaptation without any annotated data.}


Our key contribution is unifying attribute generation, model adaptation, and transductive inference within a single framework for zero- and few-shot classification. Iterative and stage-wise learning can be viewed as the optimization of a single objective, enabling both label inference within the dataset and end-to-end fine-tuning of the underlying VLMs on target domains (Alg.~\ref{algo:jtclip}). While these ideas have been explored individually, their integration is novel and leads to significant improvements over the current state-of-the-art across diverse datasets, with minimal additional computational requirements.

\section{Methodology}\label{sec:methodology}

 The input to our approach is a set of images ${\cal X} = \{\mathbf{x}_i\}_{i=1}^N$ and a set of classes ${\cal Y} = \{y_i\}_{i=1}^M$. In the zero-shot setting the goal is to assign each image to one of the $M$ classes. In the few-shot setting we are also provided with a few labeled examples ${\cal D}_{train} = \{(\mathbf{x}_i, y_i)\}_{i=1}^K$ with $\mathbf{x} \in {\cal X}_{train}$, ${\cal X}_{train} \cap {\cal X} = \O$, and $y \in {\cal Y}$. 
 
 We also consider a setting where labeled data comes from a different set of classes, i.e., $y \in {\cal Y}_{train}$ where ${\cal Y}_{train} \cap {\cal Y} = \O$. This setup is used in approaches where labeled data from a set of base categories is used to adapt CLIP on the target domain. 
 
 We report the mean per-class accuracy on the target set of images ${\cal X}$ given their ground-truth labels. To enable zero-shot learning we assume an image encoder $\theta(\cdot)$ and a text encoder $\phi(\cdot)$ such that $ \theta (\mathbf{x})^\top \phi(\mathbf{y})$ is high for image $\mathbf{x}$ and text $\mathbf{y}$ pairs that are similar. 
 We experiment with a variety of encoder pairs based on CLIP framework. In addition we assume access to a language model (e.g., Llama3 or GPT-4o) which we can query to generate attributes for each class. 

\subsection{\ours{} formulation}
\ours{} maintains a list of attributes indexed by class denoted by ${\cal A} = ({\cal A}_j)_{j=1}^M$, where ${\cal A}_j = \{\mathbf{a}_{j,k}\}_{k=1}^{n_j}$ denotes the set of text attributes for the class $j$. The number of attributes $n_j$ can vary across classes. Like the TransCLIP~\cite{transclip} formulation we maintain $\bmu = (\bmu_j)_{j=1}^M$ and $\bS = (\bS_j)_{j=1}^M$ denoting the Gaussian mixture model (GMM) mean and diagnonal variance for each class. 

In addition we maintain a matrix of softmax class assignments $\bz \in [0, 1]^{N \times M}$, where $N$ is the number of query images and $M$ is the number of classes. In other words $\bz_{i,\cdot} \in \Delta_M$ reflects the probability of assignment over all the classes, where $\Delta_M$ is the $M$-dimensional probability simplex. Given a class $j \in \Y$ the vertical slices $\bz_{\cdot, j} \in [0, 1]^N$ represents the probability that a specific query image belongs to class $j$. After inference the class label for each image $i$ can be obtained as $\argmax_j \bz_{i,j}$.

\floatname{algorithm}{Algorithm}
\renewcommand{\algorithmicrequire}{\textbf{Require:}}
\renewcommand{\algorithmicensure}{\textbf{Ensure:}}
\newcommand{\linealgocomment}[1]{\State{{\color{gray} $\triangleright$ \textit{#1}}}}

\begin{algorithm}[t]
\caption{\ours{}}
\begin{algorithmic}[1]
\Require Query images $\X$, list of classes $\Y$, list of initial attributes indexed by class $\mathcal{A}$, image encoder $\theta$, text encoder $\phi$, number of iterations $T$.
\Ensure Fine-tuned image and text encoders $\theta, \phi$, labels $\bz$, class prototypes $\bmu,\bS$, attributes indexed by class $\mathcal{A}.$
\State{$\bz,\bmu,\bS \gets 0$}
\For{$t \gets 1$ to T}
\linealgocomment{mine attributes}
\State{$\A$ $\gets$ \textsc{GenerateAttributes}($\Y, \A, \theta, \phi$)}
\linealgocomment{transductive assignment with attributes}
\State{$\bz, \bmu, \bS$ $\gets$ \textsc{Transduct}($\X, \Y, \A, \theta, \phi$)}
\linealgocomment{fine-tune image and text encoders}
\State{$\theta, \phi$ $\gets$ \textsc{Adapt}($\X, \Y, \bz, \theta, \phi$)}
\EndFor \\
\Return $\theta, \phi, \bz, \bmu, \bS, \A$
\end{algorithmic}
\label{algo:jtclip}
\end{algorithm}

\paragraph{Zero-shot Setting.} The overall objective in this formulation is:
\begin{multline}
    {\cal L}_{\text{zero-shot}}(\bz, \bmu, \bS, \by, \theta, \phi, \A) =  -\underbrace{\frac{1}{N} \sum_{i=1}^N \bz_i^\top \log (\bp_i)}_{\text{Clustering objective}} \\
     - \underbrace{\sum_{i=1}^N \sum_{j=1}^N w_{i,j} \bz_i^\top \bz_j}_{\text{Laplacian regularizer}} + \underbrace{\sum_{i=1}^N \mathbf{KL}_\lambda (\bz_i || \hat{\by}_i)}_{\text{Agreement with text}}.\label{eq:objective}
\end{multline}

The first term is a clustering objective under a Gaussian assumption for each class, and $\bp_i = (p_{i,j})_{j=1}^M \in \Delta_M$ denotes the probability over classes for the image $\bx_i$. Let $\mathbf{f}_i = \theta(\bx_i)$, then this is defined as:
\begin{equation}
    p_{i,j} \propto \det(\bS)^{-\frac{1}{2}} \exp\left(-\frac{1}{2} (\mathbf{f}_i - \bmu_j)^\top \bS^{-1}(\mathbf{f}_i - \bmu_j)\right).
\end{equation}

The second term is a Laplacian regularizer commonly seen in spectral clustering~\cite{ng2001spectral,shi2000normalized} and semi-supervised learning settings~\cite{ando2006learning, yang2022survey}. Here $w_{i,j}$ denotes the affinity between images $\bx_i$ and $\bx_j$, and this term encourages images with high affinity to have similar predictions $\bz$. We set $w_{i,j}=\max (0, \mathbf{f}_i^\top \mathbf{f}_j)$ resulting in a positive semi-definite affinity matrix $\mathbf{W} = [w_{i,j}]$ and faster optimization procedure due to a convex relaxation.

The KL divergence term ensures alignment of predictions with text and is defined as:
\begin{equation}
    \mathbf{KL}_\lambda (\bz_i || \hat{\by}_i) = \bz_i^\top \log \bz_i - \lambda \bz_i^\top \log \hat{\by}_i;~~ \lambda > 0. \label{eq:kl}
\end{equation}

The text based predictions $\hat{\by}_i$ are obtained as softmax over the mean similarity between the image and the attribute embeddings ${\cal A}_j = \{\mathbf{a}_{j,k}\}_{k=1}^{n_j}$
\begin{equation}
    \hat{\by}_{i,j} = \frac{\exp(\bar{s}_{i,j})}{\sum_{j=1}^M \exp(\bar{s}_{i,j})},\text{~where~} \bar{s}_{i,j} = \frac{1}{n_j} \sum_{k=1}^{n_j} \theta(\bx_i) \phi(\ba_{j,k}).\label{eq:image-attribute-sim}
\end{equation}

The text and vision encoders -- $\theta$ and $\phi$ output normalized and temperature-scaled features.

\paragraph{Few-shot Setting.} In the few-shot setting we can incorporate the labeled examples ${\cal D}_{train} = \{(\mathbf{x}_i, y_i)\}_{i=1}^K$ by simply setting and fixing their $\bz_i$ to the one-hot vector corresponding to the label $y_i$.

\paragraph{Zero-shot Setting with Seen Classes.} In this setting, the labeled examples ${\cal D}_{train} = \{(\mathbf{x}_i, y_i)\}_{i=1}^K$ come from a set of base classes different from the target classes, i.e., $y \in {\cal Y}_{train}$ where ${\cal Y}_{train} \cap {\cal Y} = \O$, as part of the training set. We first fine-tune CLIP using AdaptCLIPZS on the base classes, followed by transductive inference on only the target images. While this approach does not incorporate the similarity between the training and target images, it provides a straightforward comparison against prior work on adapting CLIP to target domains.

\subsection{Optimization}
The key difference between our formulation and TransCLIP is that we also update $\theta, \phi$ and $\A$. 
We initialize $\A$ with the per-class attributes in AdaptCLIPZS, which consists of prompting the LLM as:
\definecolor{quotecolor}{rgb}{0.87, 0.95, 0.96}
\begin{quote}
    \makebox[\linewidth]{%
        \colorbox{quotecolor}{%
            \hspace*{0mm} 
            \begin{minipage}{\dimexpr\linewidth+10\fboxsep\relax} 
                \fontsize{9pt}{10pt}\selectfont 
                What characteristics can be used to differentiate \texttt{[class]} from other \texttt{[domain]} based on just a photo? Provide an exhaustive list of all attributes that can be used to identify the \texttt{[domain]} uniquely. Texts should be of the form ``\texttt{[domain]} with \texttt{[attribute]}".
            \end{minipage}%
        }%
    }
\end{quote}
where \texttt{[domain]} is coarse category, \eg ``birds" for CUB~\cite{cub_dataset}, \texttt{[class]} is the common name of the category, and \texttt{[attribute]} is a specific attribute. For example, one such description is \emph{``A bird with a small, round body shape, indicative of a Baird's Sparrow."} 

The algorithm iterates between: (1) incrementally generating class-specific attributes to update $\A$ driven by pairwise confusions; (2) attribute-augmented transductive inference to estimate $\bz, \bmu, \bS$; and (3) encoder fine-tuning using the inferred $\bz$ to update the encoders $\theta$ and $\phi$. This is outlined in Algorithm~\ref{algo:jtclip} and described below. 

\paragraph{1. Generating Attributes.}\label{sec:genattr}
Our general strategy is to query large language models (LLMs) to explore the space of attributes driven by pairwise confusions. This is inspired by a long line of work on attribute discovery driven by pairwise discrimination in the computer vision literature~\cite{patterson2012sun,maji2012discovering,parikh2011interactively}. These are appended to the corresponding lists in $\A$. For a given pair of classes, we do this by prompting the LLM as:

\definecolor{quotecolor}{rgb}{0.87, 0.95, 0.96}
\begin{quote}
    \makebox[\linewidth]{%
        \colorbox{quotecolor}{%
            \hspace*{0mm} 
            \begin{minipage}{\dimexpr\linewidth+10\fboxsep\relax} 
                \fontsize{9pt}{10pt}\selectfont 
                I have a set of attributes for \texttt{[class$_1$]} as: \texttt{[attrs$_1$]}. \\
                I have a set of attributes for \texttt{[class$_2$]} as: \texttt{[attrs$_2$]}. \\[1ex]
                Provide a few additional attributes for \texttt{[class$_1$]} which can help to distinguish it from \texttt{[class$_2$]}.  \\[1ex]
                Make sure none of the attributes already given above are repeated. The texts in the attributes texts should only talk about \texttt{[class$_1$]} and should not compare it to \texttt{[class$_2$]}. 
            \end{minipage}%
        }%
    }
\end{quote}

To keep this tractable we only generate attributes for the most confused classes. We first update $\bz$ by running attribute-augmented transductive inference given the current model and set of attributes $\A$ (Step 2)\footnote{We find it beneficial to run the transductive step before invoking the generate step (see Appendix Table~\ref{tab:single-transduct})}. Then, we find the images $\bx_i$ for which the difference in the top $2$ probabilities in $\bz_{i,\cdot}$ is lower than a threshold of $\alpha$:
\begin{align*}
    \mathcal{CC} = \{(i, \{c_1, c_2\}) \hspace{.1cm}|\hspace{.1cm} 
    \bz_{i, c_1} - \bz_{i, c_2} \leq \alpha ; c1 < c2\}.
\end{align*}
Here $c_1$ and $c_2$ are the indices of the top 2 highest probabilities in $\bz_{i,\cdot}$. We then find the class pairs $\{c_1,c_2\} \in \mathcal{CC}$ which occur more than $\beta$ times.

\label{sec:transductive_method}
\paragraph{2.~Attribute-Augmented~Transductive Inference}Given the list of attributes $\A$ we can compute the text-driven labels $\hat{\by}_i$ for each class using CLIP encoders $\theta$ and $\phi$ as described in Eq.~\ref{eq:image-attribute-sim}. Optimization of $\bz, \bmu, \bS$ can be done using the same formulation of TransCLIP~\cite{transclip}. In particular they propose an iterative procedure where they optimize $\bz$ keeping $\bmu$ and $\bS$ fixed using a Majorize-Minimization procedure (similar to EM) based on a tight-linear bound on the Laplacian term. This results in efficient decoupled updates on $\bz$. 
This is followed by updates on $\bmu$ and $\bS$ keeping the remaining variables fixed using closed form updates. The algorithm converges in a few iterations and allows scaling to large datasets. We refer the reader to the details in~\cite{transclip}.

\paragraph{3. Adapting CLIP.}
We finally fine-tune CLIP encoders $\theta,\phi$ using the current set of attributes $\A$ and the inferred labels $\bz$. For each class $j$ we find the top $k$ images with the highest scores based on $\bz_{\cdot,j}$. The set of images and corresponding attributes provide a coarse form of supervision for fine-tuning. Specifically, we adopt the objective of AdaptCLIPZS~\cite{saha2024} which takes into account class-level supervision and false negative associates since multiple text-image pairs can be considered correctly aligned in a single mini-batch training. For the few-shot setting, we simply include the labeled examples to our samples.

\paragraph{Summary.} Algorithm~\ref{algo:jtclip} can be viewed as a block coordinate descent optimization of the objective in Eq.~\ref{eq:objective}. While this is straightforward for the continuous variables—such as the GMM, assignments, and encoder parameters—optimization over the space of attributes is challenging due to its inherently discrete, non-differentiable nature. Our LLM-guided exploration provides a heuristic motivated by previous work showing that attribute-augmented CLIP improves predictions, thereby improving the $\textbf{KL}$ term (if $\bz$ is accurate) in Eq.~\ref{eq:kl}. Class-confusion-guided exploration further enriches the attribute space, targeting areas where the model might benefit most. The attributes also provide a better signal for fine-tuning the CLIP to the target domain.

\section{Experiments}
\label{sec:experiments}

\paragraph{Datasets.} We evaluate \ours{} and compare to previous work on a benchmark of 12 datasets including fine-grained ones like \textbf{CUB~\cite{cub_dataset}} (200 classes), \textbf{Flowers 102~\cite{flowers_dataset}} (102 classes), \textbf{Stanford Cars~\cite{cars_dataset}} (196 classes), \textbf{FGVC Aircrafts~\cite{aircrafts_dataset}} (100 classes) and \textbf{Food101~\cite{food_dataset}} (101 classes). The benchmark also includes datasets such as  \textbf{EuroSAT~\cite{eurosat_dataset}} (10 classes), \textbf{ImageNet~\cite{imagenet_dataset}} (1000 classes), \textbf{CalTech101~\cite{caltech101_dataset}} (100 classes), \textbf{DTD~\cite{dtd_dataset}} (47 classes), \textbf{Oxford Pets~\cite{pets_dataset}} (37 classes), \textbf{Sun397~\cite{sun_dataset}} (397 classes) and \textbf{UCF101~\cite{ucf_dataset}} (101 classes). 

\paragraph{Evaluation Metrics.} For zero-shot evaluation, we assume all test images belong to the target classes, without using any labeled images. In this setting, \ours{} utilizes only the set of unlabeled test images and target categories. For the few-shot setting, we use a few labeled images per class but report accuracy on the test images across all classes, consistent with zero-shot evaluation methods. We also evaluate in the AdaptCLIPZS setting, where half of the dataset classes are considered ``seen" and the other half ``unseen." Here, the model has access to labeled examples of the seen classes and unlabeled examples from the test set of the unseen classes in a transductive setting. Final accuracy is reported on the test images of the unseen classes. This setup enables comparison with prior work that uses labeled examples from the target domain to adapt CLIP while still measuring performance on future unseen classes.

\paragraph{Implementation Details.} 
To generate attributes, we use Llama-3.1 with a maximum token length of 500. All experiments are run on a single A100 GPU. For each pair of classes, we use the prompt described in \S~\ref{sec:genattr} to generate attributes. The threshold $\alpha$ for selecting the confused images is set to 0.1, and the hyperparameter $\beta$ is adjusted so that the cumulative count $\mathcal{CC}$ includes 5\% of the most confused images. We run \ours{} for 30 iterations (i.e., $T = 30$ in \cref{algo:jtclip}) and select the top $k=8$ images per class for fine-tuning using the labels in $\mathbf{z}$. These parameters remain fixed across all datasets, and we found our approach robust to these choices within a reasonable range (see Appendix for a sensitivity analysis). 

For our experiments, we use the ViT-B/32, ViT-B/16, and ViT-L/14 architectures of OpenAI's CLIP models (Performance using CLIP models from Meta are included in the Appendix). Fine-tuning is performed with the AdamW optimizer, using betas of $(0.9, 0.98)$, an epsilon of 1E-6, and a batch size of 32. We set a learning rate of $\gamma =$ 2E-7 and weight decay of $\lambda =$ 1E-4 for the Transformer layers of the image and text encoders, and $\gamma =$ 1E-6 and $\lambda =$ 1E-4 for the final linear projection layers. All results are reported in terms of Top-1 accuracy, averaged over 3 runs.

 
 

\section{Results}\label{sec:results}
We present results for zero-shot (\S~\ref{sec:results-zero-shot}), few-shot (\S~\ref{sec:results-few-shot}), and zero-shot with seen classes (\S~\ref{sec:results-zero-shot-seen}) setting on various datasets, followed by ablation studies (\S~\ref{sec:ablation}) and a detailed analysis of our method (\S~\ref{sec:analysis}). We also provide further experiments in our Appendix, including similar performance improvements using MetaCLIP (\S~\ref{sec:additional_experiments}), sensitivity analysis (\S~\ref{sec:sensitivity}), to show robustness of our method, and detailed visualization of generated attributes (\S~\ref{section:evolution_attribute_space}).

\begin{table*}[h!]
\setlength{\belowcaptionskip}{-10pt}
\setlength{\tabcolsep}{5pt}
\begin{center}
\caption{\textbf{Zero-shot results.} Performance of CLIP, TransCLIP-ZS, and \ours{} across datasets using ViT-B/32, ViT-B/16, and ViT-L/14 architectures. \ours{} outperforms TransCLIP-ZS in all settings except for one -- UCF101 with ViT-L/14.\vspace{-8pt}}
\label{table:full_test_set_performances}
{\small
\hspace{-0.5cm}
\begin{tabular}{c@{\hskip .1cm}c@{\hskip .1cm}l@{\hskip .3cm}cccccccccccc@{\hskip .1cm}c}

& & \textbf{Method} & \rotatebox{45}{\textbf{CUB}} & \rotatebox{45}{\textbf{Aircraft}} & \rotatebox{45}{\textbf{Cars}} & \rotatebox{45}{\textbf{Flowers }} & \rotatebox{45}{\textbf{EuroSAT}} & \rotatebox{45}{\textbf{Food}} & \rotatebox{45}{\textbf{ImageNet}} & \rotatebox{45}{\textbf{Caltech}} & \rotatebox{45}{\textbf{DTD}} & \rotatebox{45}{\textbf{Pets}} & \rotatebox{45}{\textbf{SUN}} & \rotatebox{45}{\textbf{UCF}} & \rotatebox{45}{\textbf{Average}} \\
 \\[-2ex]
\cline{2-16}
 \\[-2ex]
\multirow{4}{*}[1ex]{\rotatebox{90}{\textbf{B/32}}} 
&  & CLIP & 52.33 & 19.17 & 60.33 & 66.91 & 45.01 & 80.51 & 62.06 & 91.16 & 42.67 & 87.44 & 61.95 & 62.12 & 60.97 \\
& & TransCLIP-ZS & 56.70 & 20.13 & 63.57 & 74.54 & 58.51 & 81.38 & 65.15 & 91.72 & 50.59 & 89.32 & 67.44 & 68.01 & 65.59 \\
& & \ours{} & \textbf{60.48} & \textbf{21.21} & \textbf{64.27} & \textbf{79.74} & \textbf{77.38} & \textbf{81.54} & \textbf{66.31} & \textbf{94.16} & \textbf{57.51} & \textbf{90.81} & \textbf{70.14} & \textbf{71.00} & \textbf{69.55} \\
 \\[-2ex]
\cline{2-16}
 \\[-2ex]
\multirow{4}{*}[1ex]{\rotatebox{90}{\textbf{B/16}}} 
&  & CLIP & 55.20 & 24.75 & 65.38 & 71.38 & 47.69 & 86.10 & 66.72 & 92.86 & 43.68 & 89.13 & 62.57 & 66.75 & 64.35 \\
& & TransCLIP-ZS & 62.23 & 26.88 & 68.87 & 76.17 & 65.42 & 87.15 & 70.38 & 92.86 & 50.00 & 92.34 & 68.93 & 76.34 & 69.80 \\
& & \ours{} & \textbf{66.76} & \textbf{29.31} & \textbf{72.09} & \textbf{82.05} & \textbf{76.35} & \textbf{87.38} & \textbf{71.87} & \textbf{95.46} & \textbf{58.51} & \textbf{93.43} & \textbf{73.47} & \textbf{79.06} & \textbf{73.81} \\
 \\[-2ex]
\cline{2-16}
 \\[-2ex]
\multirow{4}{*}[1ex]{\rotatebox{90}{\textbf{L/14}}} & & CLIP & 62.03 & 32.43 & 76.82 & 79.54 & 58.07 & 90.99 & 73.48 & 94.85 & 53.66 & 93.62 & 67.59 & 74.17 & 71.44 \\
&  & TransCLIP-ZS & 70.18 & 35.01 & 78.50 & 84.29 & 69.64 & 91.88 & 77.59 & 95.17 & 59.69 & 94.55 & 73.75 & {\color{BrickRed} \textbf{81.73}} & 76.00 \\
&  & \ours{} & \textbf{76.56} & \textbf{38.58} & \textbf{82.29} & \textbf{85.87} & \textbf{80.83} & \textbf{91.91} & \textbf{78.54} & \textbf{97.36} & \textbf{64.89} & \textbf{95.83} & \textbf{76.65} & 81.28 & \textbf{79.22} \\
 \\[-2ex]
\cline{2-16}
\end{tabular}
\vspace{-20pt}
}
\end{center}
\end{table*}

\subsection{Zero-Shot Performance} \label{sec:results-zero-shot}
Table~\ref{table:full_test_set_performances} shows the zero-shot performance of \ours{} compared to the CLIP~\cite{clip} and the current state-of-the-art, TransCLIP~\cite{transclip}. We report accuracy across 12 datasets using different CLIP architecture---ViT-B/32, ViT-B/16, and ViT-L/14---along with the overall average accuracy. \ours{} improves over TransCLIP by \textbf{3.96\%}, \textbf{4.01\%}, and \textbf{3.22\%}\% and over CLIP by \textbf{8.58\%}, \textbf{9.46\%}, and \textbf{7.78\%} on average using B/32, B/16, and L/14 respectively.

The highest percentage improvements are observed with ViT-B/16, though even the strongest architectures benefit from our method. Food101~\cite{food_dataset} is the most challenging, where we see a modest average improvement of \textbf{0.14\%}. In contrast, EuroSAT~\cite{eurosat_dataset} shows the greatest improvement, with the highest single-architecture boost (\textbf{18.87\%} for ViT-B/32) and the highest average improvement across architectures (\textbf{13.67\%}). \ours{} consistently outperforms both baselines in all settings except one—namely, UCF101 with ViT-L/14. These results demonstrate that our approach is broadly applicable and that reasoning over the attribute space yields significant improvements compared to transductive inference with images alone.

\subsection{Few-Shot Performance}\label{sec:results-few-shot}
Table~\ref{table:few_shot_performance} shows the few-shot performance of our approach compared to TransCLIP. We report results using the ViT-B/16 architecture with $1$-shot, $4$-shot, and $16$-shot settings denoting the number of labeled examples per class. We use the TransCLIP-FS~\cite{transclip} setting for this. Both ours and TransCLIP can incorporate labeled examples by simply setting the corresponding entries in $\bz$ to the one hot vector corresponding to their labels, as described in \S~\ref{sec:methodology}. Performance is reported on the same set of images in the zero-shot setting.

Like the zero-shot case for CLIP ViT-B/16, we find that in every setting and every choice of $k$-shot, \ours{} outperforms TransCLIP. We find an increase of \textbf{3.41\%, 3.86\%,} and \textbf{3.01\%} for $1$-shot, $4$-shot, and $16$-shot, respectively. We observe the most pronounced performance increase for $4$-shot. Trends of improvements align with the zero-shot setting. Interestingly, we find that \textbf{zero-shot \ours{} outperforms $1$-shot TransCLIP}, saving human effort, as labeling even a single example per category can be labor-intensive for certain datasets. Furthermore, we find that the gains from transduction, attribute-guided transduction with adaptation (our approach) complement labeling efforts. This flexibility is of practical value, offering end users multiple ways to improve performance on a target dataset.

\subsection{Zero-Shot Performance with Seen Classes}\label{sec:results-zero-shot-seen}
Previous work has also evaluated zero-shot learning in a setting where labeled data from a related but different set of classes is available during training, while performance is evaluated on images from unseen classes. Approaches such as CoCoOp~\cite{cocoop}, AdaptCLIPZS \cite{saha2024} and VDT \cite{clipftaself} report results by splitting a dataset’s categories in half, treating the first half as ``seen" classes to adapt their model and measuring performance on the ``unseen" second half. The results are shown in Table~\ref{table:adaptclipzs_comp}. Performance tends to be higher in this setting in comparison to the zero-shot and few-shot experiments, as only half of the classes are considered\footnote{Only the CUB dataset has lower performance as the test split is harder than the overall dataset.} and a smaller domain shift.

For a straightforward comparison, we initialize the CLIP model with the pre-trained weights from AdaptCLIPZS and report the accuracies of TransCLIP and \ours{} on the ``unseen" classes of each dataset, using the same framework as the zero-shot setting. Note that this setup does not include a transductive term between training and testing images, thus representing a lower bound on achievable performance. Despite this, we find that \ours{} outperforms prior methods across all five datasets considered: CUB~\cite{cub_dataset}, Stanford Cars~\cite{cars_dataset}, FGVC Aircraft~\cite{aircrafts_dataset}, Flowers102~\cite{flowers_dataset}, and Food101~\cite{food_dataset}.

While AdaptCLIPZS, VDT, CoOp, CoCoOp, and CLIP-A use an inductive setup, TransCLIP and \ours{} adopt a transductive approach that benefits from having test images available in advance. This results in improvements in similar vein as the zero-shot setting. However, even with domain-specific fine-tuning of CLIP with labels, transductive inference proves advantageous, and our attribute-guided approach yields further improvements--an encouraging result. The improvements over CLIP are substantial, though this setup requires more supervision than the previous settings.

\begin{table*}[h!]
\setlength{\belowcaptionskip}{-10pt}
\setlength{\tabcolsep}{5pt}
\begin{center}
{\small
\caption{\textbf{Few-shot Results.} Performance ($1$-shot, $4$-shot, and $16$-shot) of \ours{} and TransCLIP-FS across datasets using CLIP ViT-B/16 network. We find that \ours{} outperforms TransCLIP-FS in all cases.\vspace{-10pt}}
\label{table:few_shot_performance}
\hspace{-0.5cm}
\begin{tabular}{c@{\hskip .1cm}c@{\hskip .1cm}l@{\hskip .2cm}cccccccccccc@{\hskip .1cm}c}
& & \textbf{Method} & \rotatebox{45}{\textbf{CUB}} & \rotatebox{45}{\textbf{Aircraft}} & \rotatebox{45}{\textbf{Cars}} & \rotatebox{45}{\textbf{Flowers }} & \rotatebox{45}{\textbf{EuroSAT}} & \rotatebox{45}{\textbf{Food}} & \rotatebox{45}{\textbf{ImageNet}} & \rotatebox{45}{\textbf{Caltech}} & \rotatebox{45}{\textbf{DTD}} & \rotatebox{45}{\textbf{Pets}} & \rotatebox{45}{\textbf{SUN}} & \rotatebox{45}{\textbf{UCF}} & \rotatebox{45}{\textbf{Average}} \\
 \\[-2ex]
\cline{2-16}
 \\[-2ex]
\multirow{2}{*}[0ex]{\rotatebox{0}{\textbf{1}}} & & TransCLIP-FS & 65.50 & 29.84 & 70.66 & 85.10 & 71.43 & 87.83 & 69.81 & 93.18 & 51.44 & 91.81 & 70.59 & 77.82 & 72.08 \\
 & & \ours{} & \textbf{68.50} & \textbf{31.90} & \textbf{71.24} & \textbf{92.65} & \textbf{80.87} & \textbf{88.03} & \textbf{71.26} & \textbf{93.71} & \textbf{60.11} & \textbf{94.13} & \textbf{73.65} & \textbf{79.83} & \textbf{75.49} \\
 \\[-2ex]
\cline{2-16}
 \\[-2ex]
\multirow{2}{*}[0ex]{\rotatebox{0}{\textbf{4}}} & & TransCLIP-FS & 67.96 & 35.07 & 74.14 & 92.98 & 78.95 & 86.35 & 70.24 & 93.75 & 60.50 & 92.01 & 71.43 & 79.25 & 75.22 \\
 & & \ours{} & \textbf{74.01} & \textbf{38.57} & \textbf{76.75} & \textbf{96.59} & \textbf{91.03} & \textbf{86.77} & \textbf{72.76} & \textbf{94.20} & \textbf{66.76} & \textbf{92.87} & \textbf{74.66} & \textbf{83.94} & \textbf{79.08} \\
 \\[-2ex]
\cline{2-16}
 \\[-2ex]
\multirow{2}{*}[0ex]{\rotatebox{0}{\textbf{16}}} & & TransCLIP-FS & 74.24 & 38.40 & 79.56 & 94.68 & 83.35 & 86.86 & \textbf{71.89} & 94.20 & 65.47 & 92.59 & 74.81 & 81.58 & 78.14 \\
 & & \ours{} & \textbf{78.23} & \textbf{43.10} & \textbf{81.79} & \textbf{97.44} & \textbf{91.17} & \textbf{86.96} & \textbf{73.43} & \textbf{95.94} & \textbf{71.55} & \textbf{93.20} & \textbf{76.31} & \textbf{84.62} & \textbf{81.15} \\
 \\[-2ex]
\cline{2-16}
\end{tabular}
\vspace{-15pt}
}
\end{center}
\end{table*}

\definecolor{highlightrow}{rgb}{0.68, 0.85, 0.9}
\begin{table}[h!]
\caption{\textbf{Zero-shot Results with Seen Classes.} In this setting examples from a ``seen" classes are used to adapt CLIP ViT-B/16 and evaluated on ``unseen" classes. We compare both inductive and transductive approaches. Some techniques such as VDT~\cite{clipftaself} use 3:1 split as opposed to the 1:1 used by other methods on CUB so we do not include their numbers. Tranductive inference remains beneficial, and \ours{} improves over TransCLIP. \vspace{-20pt}}
\label{table:adaptclipzs_comp}

\setlength{\tabcolsep}{4pt}
\begin{center}
{\small
\begin{tabular}{clccccccc}

\textbf{Type} & \textbf{Method} & \rotatebox{45}{\textbf{CUB}} & \rotatebox{45}{\textbf{Aircraft}} & \rotatebox{45}{\textbf{Cars}} & \rotatebox{45}{\textbf{Flowers }} & \rotatebox{45}{\textbf{Food}} \\
 \\[-2ex]
\hline
 \\[-2ex]
\multirow{6}{*}[1ex]{{{Ind.}}} & CLIP & 51.91	& 36.47	& 74.94	& 77.05	& 92.49 \\
& CoOp~\cite{coop} & --- & 22.30 & 60.40 & 59.67 & 82.26 \\
& CoCoOp~\cite{cocoop} & --- & 23.71 & 73.59 & 71.75 & 91.29 \\
& CLIP-A~\cite{clip_adapter} & --- & 33.50 & 73.30 & 71.50 & 91.20 \\
& VDT~\cite{clipftaself} & --- & 33.00	& 72.90	& 75.30	& 91.20 \\
& AdaptCLIPZS & 55.63	& 40.75	& 75.78	& 81.26	& 95.08  
 \\[-2ex]
\\\hline 
 \\[-2ex]
\multirow{2}{*}[1ex]{{{Trans.}}} & TransCLIP-ZS & 61.98	& 37.37	& 78.04	& 86.45	& 95.12 \\
& \ours{} & \textbf{64.74}	& \textbf{40.99}	& \textbf{82.17}	& \textbf{89.72}	& \textbf{95.46} \\
 \\[-2ex]
\hline
\end{tabular}
\vspace{-15pt}
}
\end{center}
\end{table}


\subsection{Ablation Studies} \label{sec:ablation}
We next aim to quantify the performance contributions of each component in \cref{algo:jtclip}. In Table~\ref{table:ablation_jtclip}, we selectively disable components of our method and report the average performance over five datasets.

We find that the largest performance gain comes from combining all the components of \ours{} -- \textsc{GenerateAttributes} which corresponds to \textit{dynamic} attributes in Table~\ref{table:ablation_jtclip}, \textsc{Transduct}, and \textsc{Adapt} with an average of $\textbf{6.96\%}$ over the considered datasets. However, using \textit{dynamic} attributes without \textsc{Adapt} but with \textsc{Transduct} leads to similar performance as using \textit{static} attributes in the same scenario. This shows that fine-tuning the model is necessary to take advantage of the \textit{dynamic} attributes. \textsc{Transduct} offers an improvement of $\textbf{3.70\%}$ over baseline inductive CLIP. Adding in \textsc{Adapt} to this setting results in an improvement of $\textbf{2.50\%}$ over the strong baseline of \textsc{Transduct}. We also observe that initializing \textsc{Transduct} with \textit{static} text attributes offers a gain of $\textbf{1.34\%}$ over just using ``a photo of a \texttt{[class]}" texts. Adding only the attributes from \textsc{GenerateAttributes} to inductive CLIP offers low improvement (\textbf{1.03\%}), but when used alongside \textsc{Transduct} and \textsc{Adapt}, it increases performance.


\subsection{Class Confusion and Attribute Space}\label{sec:analysis}

Table~\ref{tab:confcounts} presents the top confused class pairs identified by our method during the first epoch on the CUB dataset. We compare these pairs with confusion counts from a linear classifier trained on the full CUB training set using labeled data. The linear classifier is trained on the entire training set of CUB using labels, and we use ground truth labels to estimate its confusion. The comparison with our method, as described in \S~\ref{sec:methodology}, reveals that 9 out of our top 10 selected confused pairs fall within the top 10\% of confused pairs identified by the linear classifier. Overall, there is strong agreement between the most confused pairs, suggesting that the class confusions identified by our approach align well with those from a fully supervised model.

\newcommand{\cmark}{{\color{OliveGreen} \ding{51}}}
\newcommand{\xmark}{{\color{BrickRed} \ding{55}}}
\begin{table}
\small
\centering
\caption{\textbf{Ablation Study.} Ablation study of the components of \ours{} using ViT-B/16. Average Top-1 accuracy across five datasets is shown (see Appendix for the full table). Attributes $\A = \{$\O$, \text{S}, \text{D}\}$ refer to \textit{no}, \textit{static}, and \textit{dynamic} attributes, respectively. No attributes corresponds to standard CLIP, while static and dynamic refer to the initial set of attributes and confusion-driven attributes respectively. The first row shows the performance of CLIP, and the third row shows the performance of TransCLIP. Simply generating attributes leads to insignificant improvement in performance on these fine-grained datasets (row 2), but it improves transductive inference and subsequent adaptation. Dynamic attribute generation provides additional benefits (last row).}
\setlength{\tabcolsep}{4pt}
\begin{tabular}{@{\hskip .3cm}ccccccc@{}}
\toprule
 \textsc{Attributes} & \textsc{Transduct} & \textsc{Adapt} & \textbf{Acc.} & $\Delta$ \textbf{CLIP} \\
\\[-2ex]
\hline
 \\[-2ex]
 $\O$ & \xmark & \xmark & 60.56 & --- \\
 S & \xmark & \xmark & 61.59 & {\color{OliveGreen} +1.03\%} \\
 $\O$ & \cmark & \xmark & 64.26 & {\color{OliveGreen} +3.70\%} \\
S&  \cmark & \xmark & 65.60  & {\color{OliveGreen} +5.04\%} \\
S & \cmark & \cmark & 66.76 & {\color{OliveGreen} +6.20\%} \\
D&  \cmark & \xmark & 65.56  & {\color{OliveGreen} +5.00\%} \\
D & \cmark & \cmark & \textbf{67.52} & {\color{OliveGreen} \textbf{+6.96\%}}\\
\\[-2ex]
\hline
\end{tabular}
\vspace{-10pt}
\label{table:ablation_jtclip}
\end{table}

Table~\ref{tab:confcounts} also visualizes the progression of confusion counts for the top confused pair (Western Gull, California Gull). The number of images with a probability difference below $\alpha = 0.1$ generally decreases over epochs. The most significant drop occurs between the first and second epochs, highlighting the impact of the newly generated attributes.

\begin{figure}
    \centering
    \captionsetup{type=figure}
    \includegraphics[width=0.75\linewidth]{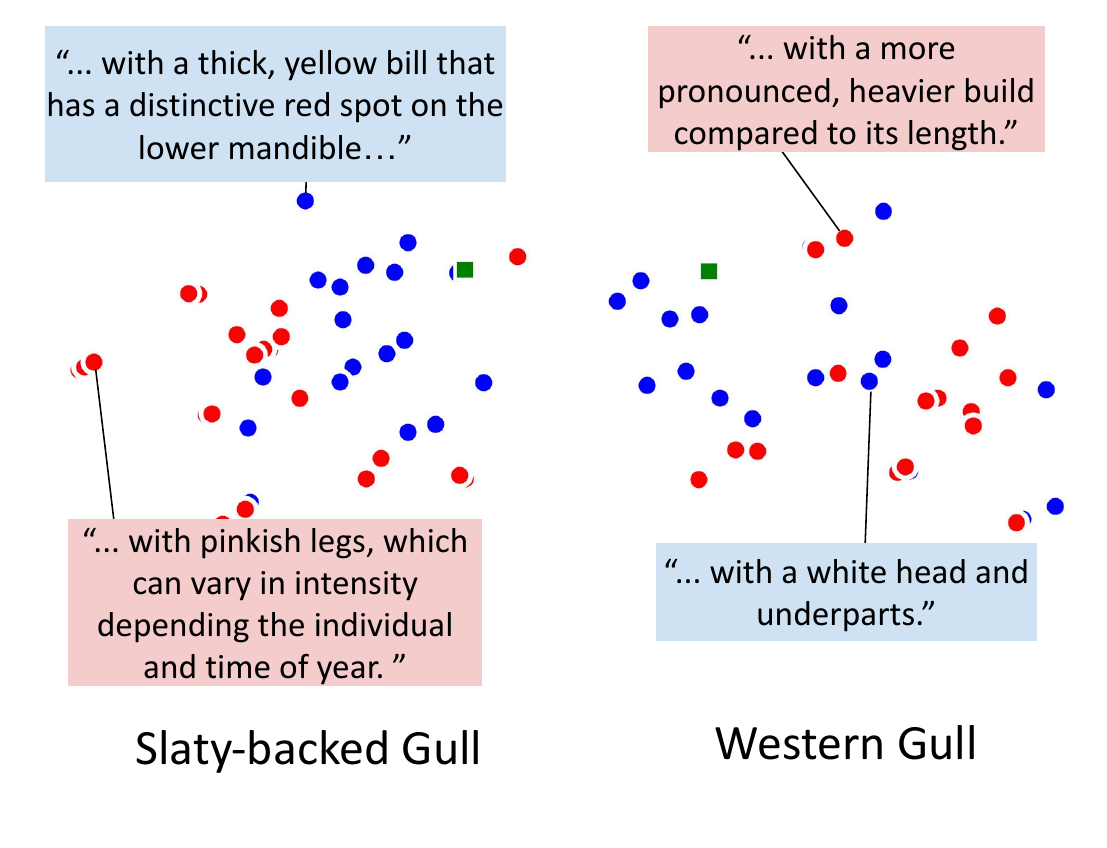}
    \vspace{-10pt}
    \captionof{figure}{\textbf{t-SNE Plots of Class Attributes.} For each category the prototype, initial set of attributes, and the final set of attributes are shown in green, blue, and red respectively. Habitat, relative characteristics, and other distinguishing features are often identified through pairwise comparisons, while the initial attributes tend to describe the prominent visual features. These plots were obtained by mapping the CLIP text embeddings of the attributes using t-SNE. Please see the Appendix for detailed figures.}
    \label{fig:tsne_attributes}
    \vspace{-10pt}
\end{figure}

Figure~\ref{fig:tsne_attributes} illustrates the evolution of the attribute space across various categories. For each category, the class prototype (``photo of [class]"), the initial set of attributes, and the final set of attributes are shown in green, blue, and red, respectively. These visualizations were generated by projecting CLIP text embeddings of the attributes using t-SNE. Several new attributes were added through pairwise comparisons, and a few notable examples are highlighted in the figure. Many of the attributes discovered through pairwise comparison highlight differences in habitat, relative characteristics (e.g., “...more pronounced build compared to its length” for the Western Gull), and other distinguishing features. Bird images often include backgrounds indicative of habitat types, and this form of supervision enables CLIP to learn to associate these attributes with categorization. Larger versions of these figures are in the Appendix.

\begin{table}[t]
    \centering
        \caption{\textbf{Class Confusions.} (Left) Progressive reduction of pairwise confusion between ``Western Gull" and ``California Gull" over training iterations of \ours{}. (Right) Most confused class pairs according to our selection criteria. We show the counts of pairwise misclassified test images according to our procedure and according to a linear classifier trained on the labeled training images using the CLIP image features.}
        \label{tab:confcounts}
    \begin{minipage}{0.28\columnwidth} 
        \centering
        \includegraphics[width=\linewidth]{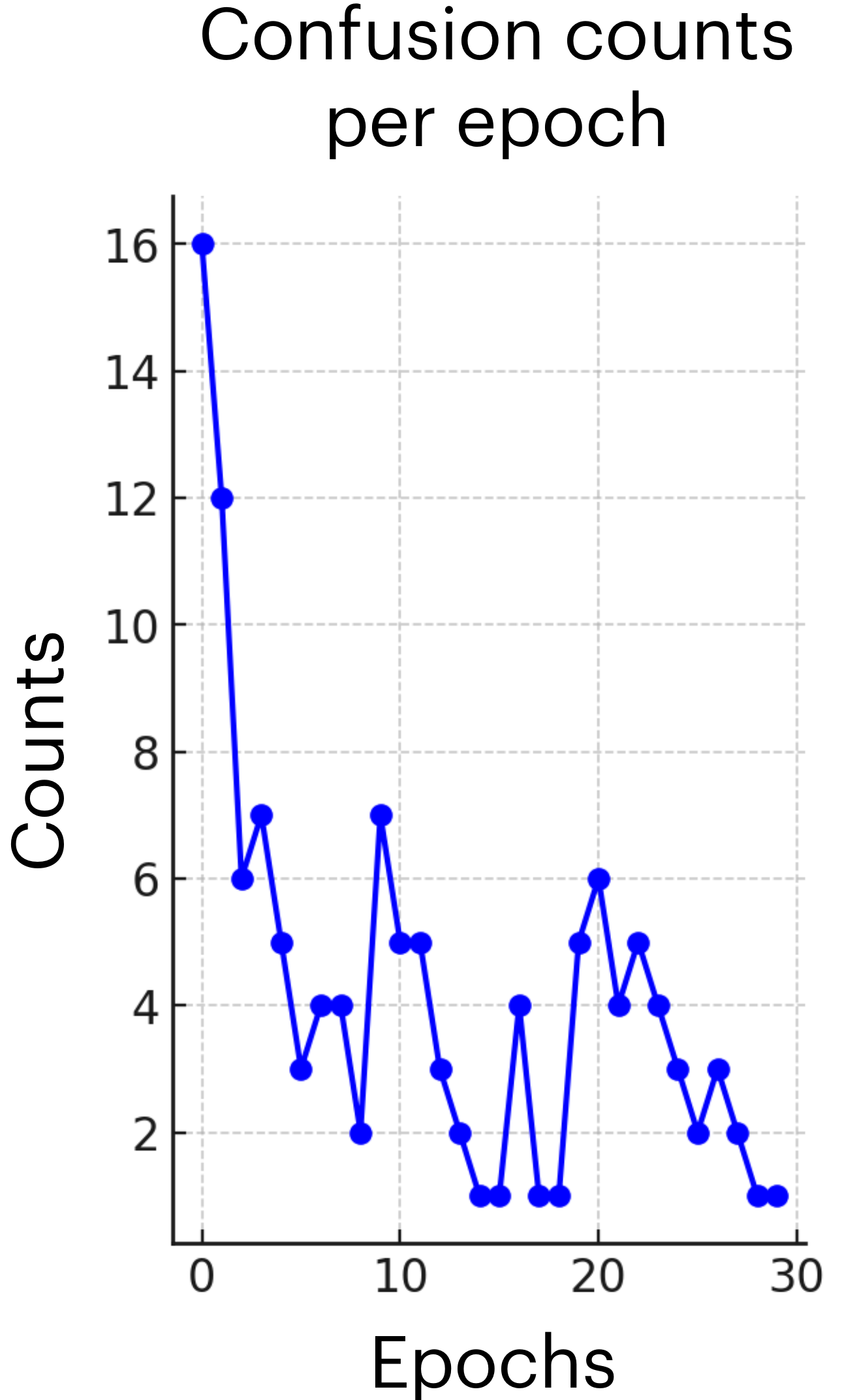}
    \end{minipage}
    \hfill
    \begin{minipage}{0.68\columnwidth} 
        \centering
        \setlength{\tabcolsep}{3pt}
        \scriptsize
        \begin{tabular}{@{\hskip -1pt}r@{\hskip 2pt}l@{\hskip 2pt}c@{\hskip 3pt}c@{\hskip -1pt}}
            \hline
            \multicolumn{2}{c}{\textbf{Confused Class Pairs}} & Ours & \cellcolor[HTML]{FFFFFF}Linear \\
            \hline
            California Gull & Western Gull & 16 & \cellcolor[HTML]{FFFFFF}10 \\
            Least Flycatcher & Olive sided Flycatcher & 13 & \cellcolor[HTML]{FFFFFF}3 \\
            American Crow & Common Raven & 8 & \cellcolor[HTML]{FFFFFF}12 \\
            Least Flycatcher & Western Wood Pewee & 7 & \cellcolor[HTML]{FFFFFF}6 \\
            Eared Grebe & Horned Grebe & 7 & \cellcolor[HTML]{FFFFFF}11 \\
            Bronzed Cowbird & Shiny Cowbird & 6 & \cellcolor[HTML]{FFFFFF}3 \\
            Brewer Sparrow & Harris Sparrow & 6 & \cellcolor[HTML]{FFFFFF}2 \\
            Slaty backed Gull & Western Gull & 5 & \cellcolor[HTML]{FFFFFF}6 \\
            Baird Sparrow & Grasshopper Sparrow & 5 & \cellcolor[HTML]{FFFFFF}3 \\
            Philadelphia Vireo & Warbling Vireo & 5 & \cellcolor[HTML]{FFFFFF}8 \\
            \hline
        \end{tabular}
    \end{minipage}
\end{table}

\subsection{Computational Cost}\label{sec:computation}
For attribute expansion, we explore both open-source Llama-3.1-8b and GPT-4o, and find their effect on performance to be similar (see Table~\ref{tab:gpt4o} in Appendix). The number of class pairs requiring attribute expansion decreases with each epoch, reaching zero for most datasets after about 10 epochs due to the chosen thresholds $\alpha$ and $\beta$, ensuring that attributes are not regenerated for duplicate class pairs. 

In the CUB dataset, a total of approximately 30 pairs of confusing classes were selected over 30 iterations for prompting the LLM. Across datasets of various sizes, the number of sampled class pairs remains within a similar range, with the Flowers dataset requiring attribute expansion for only three pairs. For larger datasets such as ImageNet, we lower $\alpha$ to 0.05 to keep attribute expansion computationally feasible. Running Llama-3.1-8b on our hardware (a single A100) takes less than 10 minutes for all selected pairs, while using GPT-4o via API calls takes under 2 minutes and costs less than \$1. 

Our fine-tuning process is highly efficient, using only 8 examples per category. For a dataset like CUB, which contains approximately 12k images across 200 classes, 30 iterations of fine-tuning and transduction take less than 10 minutes on a single A100 GPU.

Overall, the runtime cost of \ours{} is approximately 10-20 minutes higher than that of CLIP on most datasets. However, these additional costs can be justified given the performance gains, since manually labeling even a small fraction of the dataset would require significantly more time, such as in fine-grained domains.


\section{Limitations}
There are two main limitations to our work. The first is the use of LLMs to generate fine-grained attributes. LLMs are known to hallucinate data, posing a risk of generating incorrect attributes. However, similar to AdaptCLIPZS~\cite{saha2024}, which demonstrated through human evaluation that the generated attributes are highly accurate, we did not find this to be a significant issue for the datasets we considered. Additionally, this step can be guided by domain experts.

The second limitation is the applicability of the transductive setup. It requires access to the entire test set at once, which may not always be feasible (e.g., in a streaming setting). Additionally, we have not demonstrated robustness to imperfect knowledge of class distributions and data structure. For instance, we assume that images within a class cluster together, allowing us to model each class as a Gaussian distribution, which may not always be true.

\section{Conclusion}
\label{sec:conclusion}
The transductive setting offers a compelling approach for practitioners and domain experts who need precise answers for specific datasets. For instance, an ecologist might be interested in estimating species counts from data gathered via a network of camera traps, while a scientist might want to determine land-cover distribution using satellite imagery. VLMs enable straightforward labeling through language-based descriptions of categories, but their initial accuracy is often insufficient. Our work demonstrates that expanding categories based on attributes, when combined with transductive learning, enables model fine-tuning to achieve significant accuracy improvements. Additionally, this approach offers complementary advantages to traditional labeling methods, such as providing a few labeled examples per class. While we use large language models for convenience, this iterative procedure is naturally suited to human-in-the-loop approaches, allowing practitioners to incrementally add attributes and labels for ambiguous classes. These findings are practically valuable, as they offer end-users multiple pathways to improve labeling precision on their target dataset without investing significant efforts on training dataset curation and model training.

\section{Acknowledgements}
The research is supported in part by grant \#2329927 from the National Science Foundation (USA). Our experiments were performed on the GPU cluster funded by the Mass. Technology Collaborative.

{
    \small
    \bibliographystyle{ieeenat_fullname}
    \bibliography{main}
}

\clearpage

\maketitlesupplementary
\setcounter{page}{1}

\section{Additional Ablations}
\label{sec:additional_experiments}

We explore additional ablative studies over \ours{} and its various components in this section.

\begin{table*}[!hb]
\small
\centering
\caption{\textbf{Per-dataset results of Ablation Study.} For five datasets on the ViT-B/16 architecture, we present the effect of various components of \ours{}. We use the same conventions as Table~\ref{table:ablation_jtclip}.}
\begin{tabular}{ccccccccc}
\toprule
\textsc{Attributes} & \textsc{Transduct} & \textsc{Adapt} & \multicolumn{1}{l}{CUB} & \multicolumn{1}{l}{Cars} & \multicolumn{1}{l}{Aircraft} & \multicolumn{1}{l}{Flower} & \multicolumn{1}{l}{Food} & \multicolumn{1}{l}{\textbf{Average}} \\\\[-2ex]
\hline
 \\[-2ex]
$\O$ & \xmark & \xmark & 55.20 & 65.38 & 24.75 & 71.38 & 86.10 & 60.56 \\
S & \xmark & \xmark & 57.70 & 65.65 & 24.78 & 73.33 & 86.50 & 61.59 \\
$\O$ & \cmark & \xmark & 62.23 & 68.87 & 26.88 & 76.17 & 87.15 & 64.26 \\
S & \cmark & \xmark & \cellcolor[HTML]{FFFFFF}64.15 & \cellcolor[HTML]{FFFFFF}69.83 & \cellcolor[HTML]{FFFFFF}26.73 & \cellcolor[HTML]{FFFFFF}80.06 & \cellcolor[HTML]{FFFFFF}87.25 & 65.60 \\
S & \cmark & \cmark & 65.86 & 71.33 & 28.62 & 80.67 & 87.30 & 66.76 \\
D & \cmark & \xmark & 64.20 & 69.53 & 26.58 & 80.23 & 87.27 & 65.56 \\
D & \cmark & \cmark & \textbf{66.76} & \textbf{72.09} & \textbf{29.31} & \textbf{82.05} & \textbf{87.38} & \textbf{67.52}\\
 \\[-2ex]
\hline
\end{tabular}
\label{tab:ablationperdataset}
\end{table*}

\begin{table*}[!htbp]
\centering
\caption{\textbf{Performance with MetaCLIP.} We change the base VLM from CLIP to MetaCLIP for TransCLIP and \ours{} and observe consistent improvements over the baselines on ViT-B/16}
\begin{tabular}{lcccccc}
\toprule
Method & CUB & Cars & Aircraft & Flower & Food & \textbf{Average} \\
 \\[-2ex]
\hline
 \\[-2ex]
CLIP & 55.20 & 65.38 & 24.75 & 71.38 & 86.10 & 60.56 \\
TransCLIP & 62.23 & 68.87 & 26.88 & 76.17 & 87.15 & 64.26 \\
\ours{} & \textbf{66.76} & \textbf{72.09} & \textbf{29.31} & \textbf{82.05} & \textbf{87.38} & \textbf{67.52}\\
 \\[-2ex]
\hline
 \\[-2ex]
MetaCLIP & 68.67 & 74.49 & 28.65 & 73.81 & 84.01 & 65.93 \\
TransMetaCLIP & 74.02 & 79.01 & 31.56 & 80.15 & 85.53 & 70.05 \\
GTA-MetaCLIP & \textbf{78.36} & \textbf{82.30} & \textbf{35.58} & \textbf{81.57} & \textbf{85.98} & \textbf{72.76}\\
 \\[-2ex]
\hline
\end{tabular}
\label{tab:metaclip}
\end{table*}

\begin{table*}[!htbp]
\centering
\caption{\textbf{Effect of LLM model on accuracy of \ours{}.} We switch the LLM model used by \textsc{GenerateAttributes} from Llama-3.1 to GPT4o and observe similar performance on ViT-B/16.}
\begin{tabular}{lcccccc}
\toprule
LLM & CUB & Cars & Aircraft & Flower & Food & \textbf{Average} \\
 \\[-2ex]
\hline
 \\[-2ex]
GPT4o & 66.50 & 72.13 & 29.89 & 81.55 & 87.36 & 67.49 \\
Llama-3.1 & 66.76 & 72.09 & 29.31 & 82.05 & 87.38 & 67.52\\
 \\[-2ex]
\hline
\end{tabular}
\label{tab:gpt4o}
\end{table*}

\begin{table*}[!htbp]
\centering
\caption{\textbf{Removing the internal transductive update step in \textsc{GenerateAttributes}}, thereby making only a single call to \textsc{Transduct} per iteration reduces the accuracy on average over five datasets on the ViT-B/16 architecture.}
\begin{tabular}{lcccccc}
\toprule
LLM & CUB & Cars & Aircraft & Flower & Food & \textbf{Average} \\
 \\[-2ex]
\hline
 \\[-2ex]
\ours{} \textit{single \textsc{Transduct}} & 66.72 & 69.89 & 29.55 & 81.32 & 87.32 & 66.96 \\
\ours{} \textit{original} & 66.76 & 72.09 & 29.31 & 82.05 & 87.38 & 67.52\\
 \\[-2ex]
\hline
\end{tabular}
\label{tab:single-transduct}
\end{table*}

\definecolor{highlightrow}{rgb}{0.68, 0.85, 0.9}
\begin{table*}[!h]
\setlength{\belowcaptionskip}{-1pt}
\setlength{\tabcolsep}{7pt}
\begin{center}
\caption{\textbf{Sensitivity analysis over the top-$k$ and $T$ selection} of \ours{} using the CLIP ViT-B/16 architecture without the dynamic \textsc{GenerateAttributes} component (ie. TransCLIP$^{FT}$ in \cref{table:full_test_set_performances}) as given in \cref{algo:jtclip}. We pick $k=8,T=30$ even though there exist better performing alternatives. We fix this hyperparameter selection to ablate on the remaining parameters of \ours{}.}
\label{table:ablation_jtclip_topk_T}
{
\begin{tabular}{l@{\hskip 7pt}lcccccc}
\hline
 \\[-2ex]
\textbf{top-$k$} & \textbf{$T$} & \rotatebox{90}{\textbf{CUB \cite{cub_dataset}}} & \rotatebox{90}{\textbf{Cars \cite{cars_dataset}}} & \rotatebox{90}{\textbf{Aircrafts \cite{aircrafts_dataset}}} & \rotatebox{90}{\textbf{Flowers \cite{flowers_dataset}}} & \rotatebox{90}{\textbf{Food \cite{food_dataset}}} & \rotatebox{90}{\textbf{Average}} \\[.4ex]
\\[-2ex]
\hline
 \\[-2ex]
1 & 30 & 65.93 & 71.55 & \textbf{29.43} & 81.04 & 87.36 & 67.06 \\
3 & 30 & 65.64 & \textbf{71.97} & 28.95 & \textbf{82.01} & \textbf{87.43} & \textbf{67.20} \\
5 & 30 & 65.64 & \textbf{71.97} & 28.74 & 81.28 & 87.39 & 67.00 \\
\rowcolor{highlightrow} 8 & 30 & 65.86 & 71.33 & 28.62 & 80.67 & 87.30 & 66.76 \\
10 & 30 & 65.84 & 71.45 & 28.53 & 81.04 & \textbf{87.43} & 66.86 \\
\textbf{20} & 30 & \textbf{66.09} & \textbf{71.97} & 28.29 & 81.04 & 87.36 & 66.95 \\
 \\[-2ex]
\hline
 \\[-2ex]
8 & 1 & 63.98 & 69.87 & 27.48 & 80.76 & 87.37 & 65.89 \\
8 & 10 & 65.05 & 70.55 & 28.17 & 81.04 & 87.36 & 66.43 \\
8 & 20 & 65.48 & 71.11 & 28.47 & 81.04 & \textbf{87.43} & 66.71 \\
\rowcolor{highlightrow} 8 & 30 & 65.86 & 71.33 & 28.62 & 80.67 & 87.30 & 66.76 \\
8 & 40 & \textbf{66.14} & 72.63 & 28.80 & 81.04 & 87.34 & 67.19 \\
8 & 50 & \textbf{66.14} & \textbf{72.71} & \textbf{28.98} & \textbf{82.01} & \textbf{87.43} & \textbf{67.46} \\
 \\[-2ex]
\hline
\end{tabular}
\vspace{.3cm}
}
\end{center}
\end{table*}

\definecolor{highlightrow}{rgb}{0.68, 0.85, 0.9}
\begin{table*}
\setlength{\belowcaptionskip}{-10pt}
\setlength{\tabcolsep}{9pt}
\begin{center}
\caption{\textbf{Ablation over the probability threshold $\alpha$} of the \textsc{GenerateAttributes} implementation of \ours{} as given in \cref{sec:experiments} using $k = 8, T = 30$ as determiend from \cref{table:ablation_jtclip_topk_T}. Like \cref{table:ablation_jtclip_topk_T}, even though there are better performing selection, we choose $\alpha = 10\%$.}
\label{table:ablation_jtclip_generate_thresholds}
{
\begin{tabular}{lcccccc}
\hline
 \\[-2ex]
\textbf{$\alpha$} & \rotatebox{90}{\textbf{CUB \cite{cub_dataset}}} & \rotatebox{90}{\textbf{Cars \cite{cars_dataset}}} & \rotatebox{90}{\textbf{Aircrafts \cite{aircrafts_dataset}}} & \rotatebox{90}{\textbf{Flowers \cite{flowers_dataset}}} & \rotatebox{90}{\textbf{Food \cite{food_dataset}}} & \rotatebox{90}{\textbf{Average}}   \\[.4ex]
\\[-2ex]
\hline
 \\[-2ex]
2.5\% & 65.67 & 71.68 & 28.50 & \textbf{83.23} & 87.27 & 67.27 \\
5.0\% & 66.69 & 71.67 & 28.50 & 81.04 & 87.40 & 67.06 \\
7.5\% & \textbf{66.98} & 71.56 & 28.98 & 80.67 & 87.32 & 67.10 \\
\rowcolor{highlightrow} 10.0\% & 66.76 & 72.09 & \textbf{29.31} & 82.05 & 87.38 & 67.52 \\
12.5\% & 65.48 & 72.64 & 28.47 & 82.01 & \textbf{87.50} & 67.22 \\
15.0\% & 66.90 & 71.74 & 28.65 & 82.05 & 87.41 & 67.35 \\
17.5\% & 66.83 & 72.65 & 28.83 & 82.42 & 87.29 & \textbf{67.61} \\
20.0\% & 66.72 & \textbf{72.99} & 28.95 & 80.88 & 87.33 & 67.37 \\
 \\[-2ex]
\hline
\end{tabular}
}
\end{center}
\end{table*}

\paragraph{Per Dataset Results.} Table~\ref{tab:ablationperdataset} breaks down the Top-1 accuracies across reported in Table~\ref{table:ablation_jtclip} of the main papers across individual datasets namely CUB, Stanford Cars, FGVC Aircraft, Flowers102, and Food101 datasets using the ViT-B/16 architecture. We observe similar trends for each dataset for all the ablations considered.

\paragraph{Using MetaCLIP as the base VLM.} MetaCLIP introduces better CLIP architectures by curating training data and scaling training. We switch the base VLM from CLIP to MetaCLIP to take advantage of this and test the generalization of our approach to new architectures. Table~\ref{tab:metaclip} presents the accuracies for the inductive version of MetaCLIP, TransMetaCLIP (the TransCLIP method applied to MetaCLIP), and GTA-MetaCLIP (our method applied to MetaCLIP). The experiments are conducted using the ViT-B/16 architecture of MetaCLIP across CUB, Stanford Cars, FGVC Aircraft, Flowers102, and Food101 datasets. We observe consistent improvements in the case of MetaCLIP too. On average, over the five datasets, we see an improvement of 6.8\% over MetaCLIP, and an improvement of 2.7\% over TransMetaCLIP on using our method. This is similar to our improvements of 7.0\% and 3.3\% on the corresponding baselines with CLIP.

\paragraph{Effect of the LLM Model in \textsc{GenerateAttributes}.} For all the results in the main paper, we used Llama-3.1 as the LLM model for dynamic attribute generation. Now we explore using GPT4o as the LLM in Table~\ref{tab:gpt4o}. We observe that the accuracy remains similar on average over the CUB, Stanford Cars, FGVC Aircraft, Flowers102, and Food101 datasets on ViT-B/16. Thus, using Llama-3.1 is a more cost-effective choice for dynamic attribute generation due to its open-source nature.

\textbf{Removing the internal call to \textsc{Transduct} in \textsc{GenerateAttributes}.} 
\label{sec:singletransduct}
We remove the internal transductive inference update call (see \S~\ref{sec:genattr}) in \textsc{GenerateAttributes} and present the results over five datasets using the ViT-B/16 architecture in Table~\ref{tab:single-transduct}. We observe that on average the accuracy drops on removing this call to \textsc{Transduct}. For Aircraft, we observe that the accuracy slightly improves on dropping this \textsc{Transduct} call, however for Cars we see a significant decrease.

\section{Sensitivity Analysis}
\paragraph{Top-$k$ Selection and Number of Iterations $T$.} In \cref{table:ablation_jtclip_topk_T} we show the performance of \ours{} when varying top-$k$ and $T$ selection. The table is divided into two sections: first we fix $T$ and sweep over $k$, and secondly we fix $k$ and sweep over $T$. We find that increasing $T$ has the strongest correlation with performance, with average performance across benchmarks monotonically increasing for $k = 8$ when going from $T = 1$ to $T = 50$. Furthermore, we find that Flowers and Food are the most insensitive to changes in hyperparameters, keeping mostly the same value irrespective of $k$ and $T$. Overall, we find the performance guarantees to be quite high even in the worst case ($65.89$ with $k = 8, T = 1$), still being higher than default TransCLIP ($64.26$) or TransCLIP with static fine-grained attributes ($65.60$).

\paragraph{Probability Threshold $\alpha$.} Similarly, in \cref{table:ablation_jtclip_generate_thresholds} we show the performance of \ours{} when varying the probability threshold for determining confusing pairs of classes, $\alpha$. For the whole experiment, we fix $k = 8, T = 30$ and sweep over $alpha$. We find that each benchmark has it's own ideal $\alpha$ value, namely that no two benchmark's max performances share an common $alpha$. Surprisingly, we see that $\alpha = 17.5\%$, which does not perform the best on any benchmark, has the highest average value. We also conclude that \ours{} has a greater insensitivity to the choice of $\alpha$ as compared to $T$ but similar to $k$. Namely we find that the spread of $\alpha$ to be $67.61 - 67.06 = 0.55$, $T$ to be $67.46 - 65.89 = 1.57$, and $k$ to be $67.20 - 66.76 = 0.44$. Finally, we find that the minimum performance increase by introducing \textsc{GenerateAttributes} is at $\alpha = 5.0\%$ with a gain of $67.10 - 66.76 = 0.34$. In other words, adding any amount of comparative attribute generation improves performance.

\label{sec:sensitivity}

\section{Evolution of Attribute Space}
\label{section:evolution_attribute_space}
\begin{figure*}
    \centering
    \captionsetup{type=figure}
    \includegraphics[width=.95\linewidth]{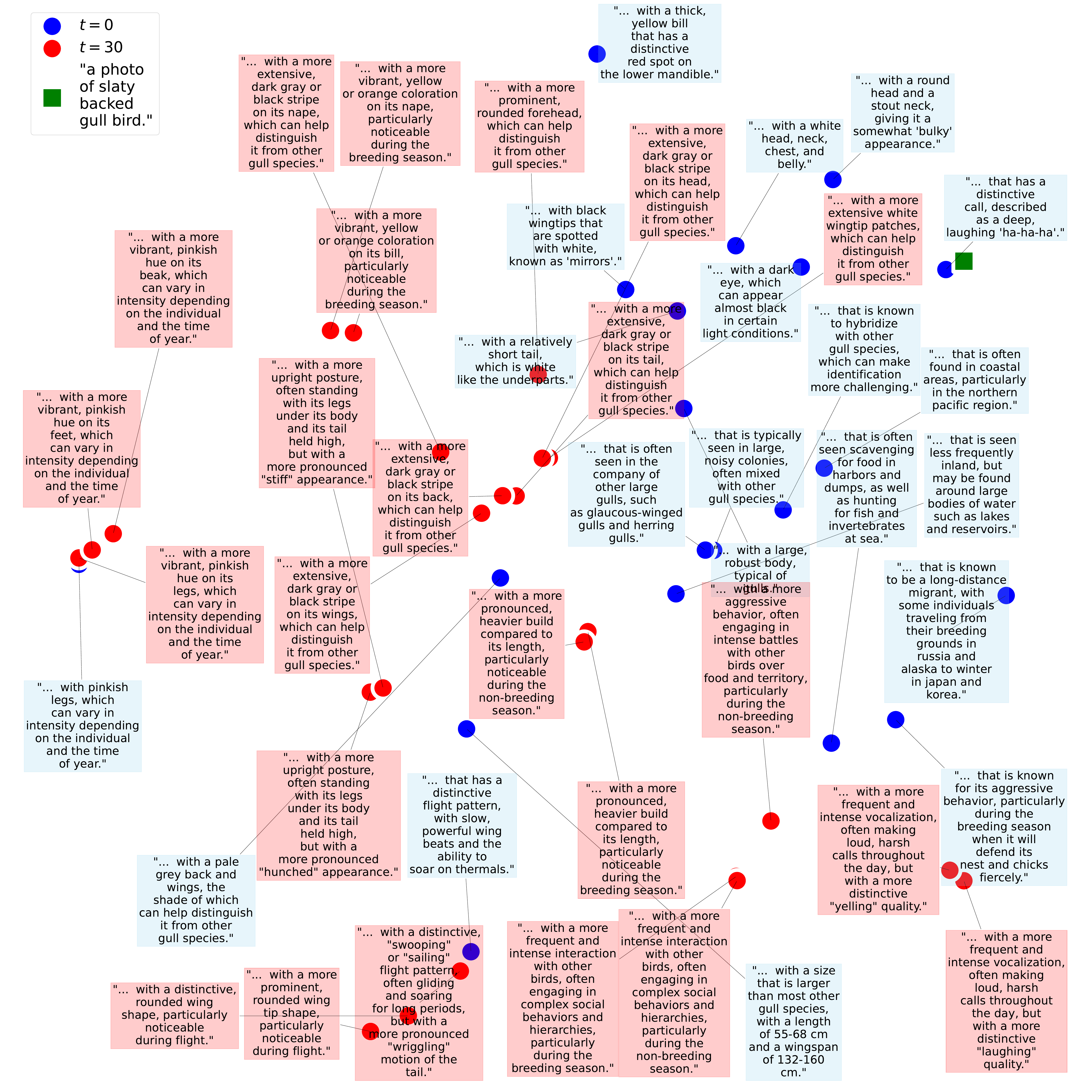}    \captionof{figure}{\textbf{Slaty-backed Gull (vs. Western Gull) Annotated T-SNE Plot.}}
    \vspace{-.2cm}
    \label{fig:tsne_appendix_1}
\end{figure*}

\begin{figure*}
    \centering
    \captionsetup{type=figure}
    \includegraphics[width=.95\linewidth]{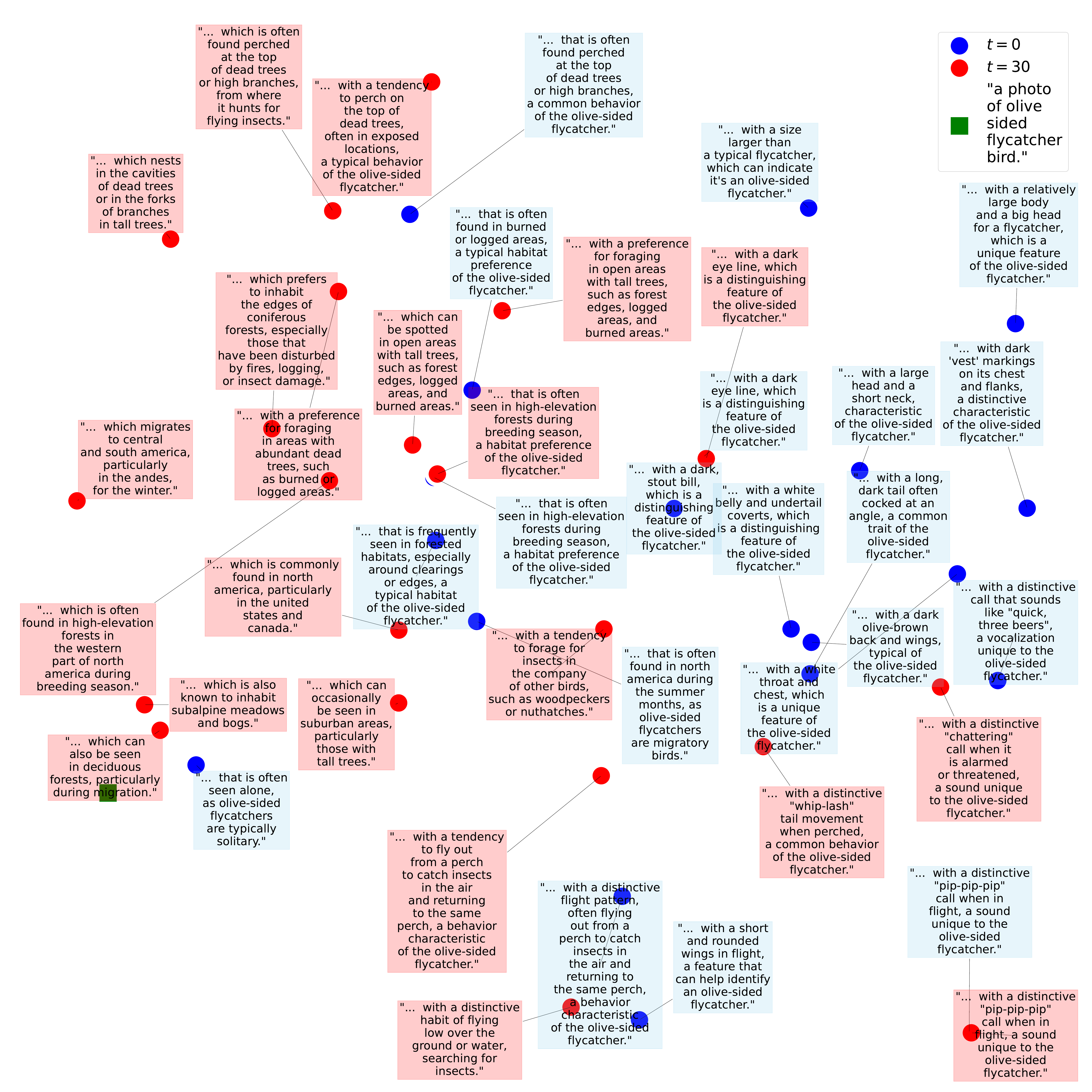}    \captionof{figure}{\textbf{Olive-sided Flycatcher (vs. Least Flycatcher) Annotated t-SNE Plot.}}
    \vspace{-.2cm}
    \label{fig:tsne_appendix_2}
\end{figure*}

\begin{figure*}
    \centering
    \captionsetup{type=figure}
    \includegraphics[width=.95\linewidth]{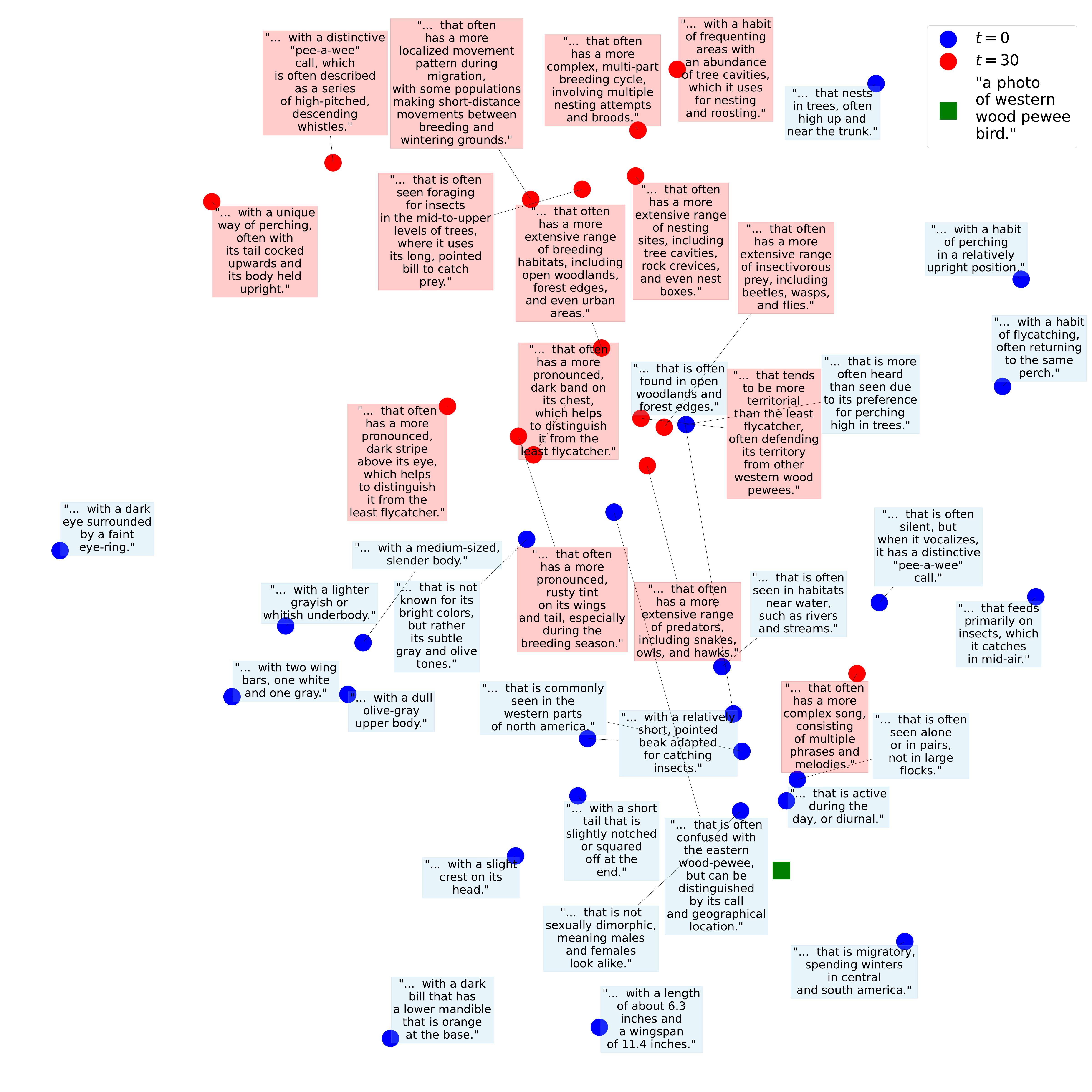}    \captionof{figure}{\textbf{Western Wood-Pewee (vs. Least Flycatcher) Annotated t-SNE Plot.}}
    \vspace{-.2cm}
    \label{fig:tsne_appendix_3}
\end{figure*}

\begin{figure*}
    \centering
    \captionsetup{type=figure}
    \includegraphics[width=.95\linewidth]{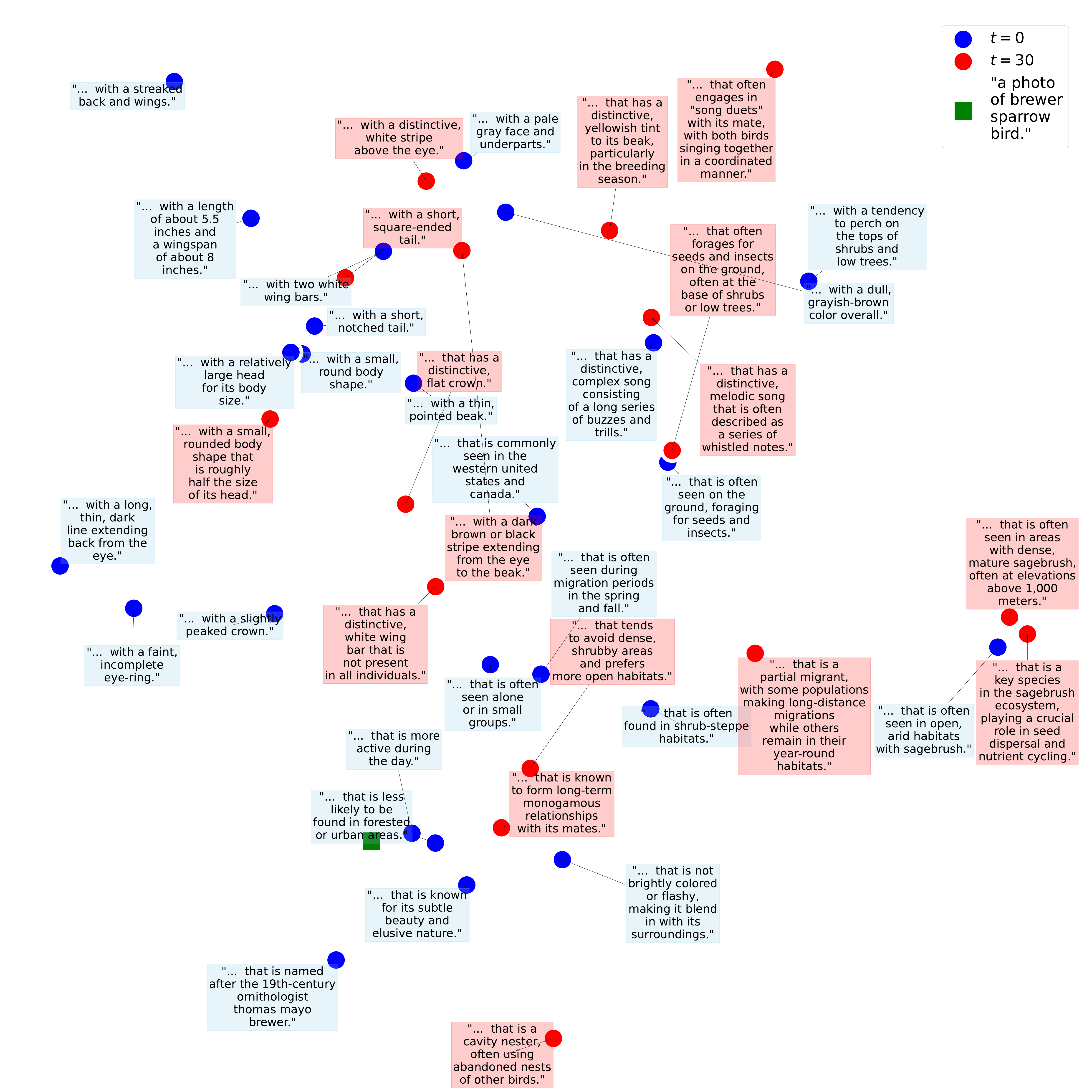}    \captionof{figure}{\textbf{Brewer's Sparrow (vs. Harris' Sparrow) Annotated t-SNE Plot.}}
    \vspace{-.2cm}
    \label{fig:tsne_appendix_4}
\end{figure*}

\begin{figure*}
    \centering
    \captionsetup{type=figure}
    \includegraphics[width=.95\linewidth]{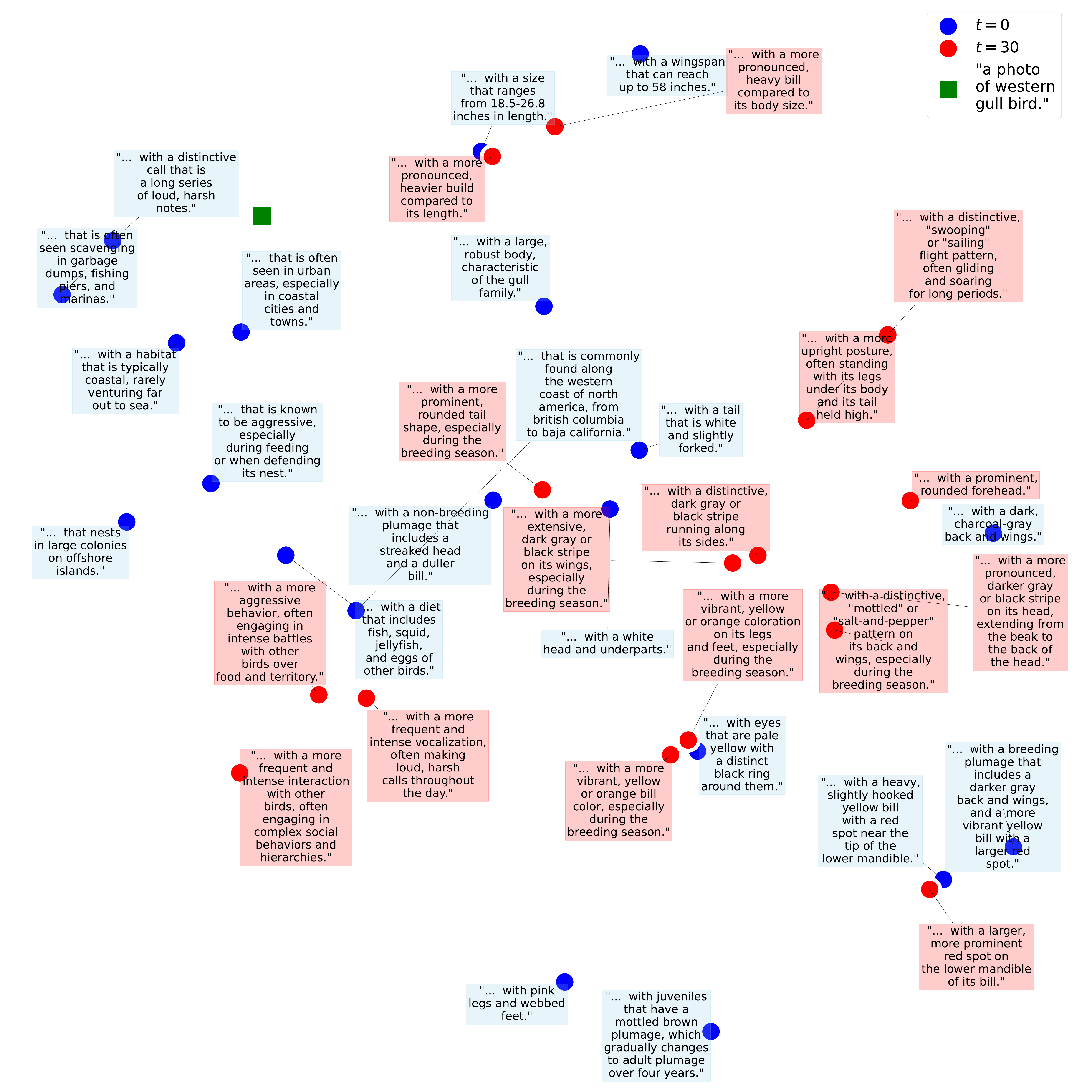}    \captionof{figure}{\textbf{Western Gull (vs. California Gull) Annotated t-SNE Plot.}}
    \label{fig:tsne_appendix_5}
\end{figure*}

In \cref{fig:tsne_appendix_1} through \cref{fig:tsne_appendix_5}, we depict the evolution of the set of attributes for a given class over the course of our method. \ours{} begins with a list of static fine-grained attributes (depicted in blue) and through iterations of the method generates additional comparative attributes between confusing classes (red). We embed these attributes with the CLIP text tower and use t-SNE to visualize the relative locations of these attributes. The specific prompt generated for a given point is indicated within the figure. We see that attributes within the reduced space often form tight clusters grouped by similar concepts (eg. "habitat" or "appearance"). When dynamically generated attributes (red points) are close to the initial static attributes (blue) we see more similar semantic meaning. Finally, one can notice that the newly added attributes occupy different regions of the space, namely that using dynamic generation effectively expands the list of fine-grained details on a given class.

\newpage


\end{document}